\documentclass[letterpaper, 10 pt, conference]{ieeeconf}  
\IEEEoverridecommandlockouts                              
\usepackage{graphics,graphicx}           
\usepackage{amsmath,amssymb}    
\usepackage{newtxtext,newtxmath}
\usepackage{bm}
\usepackage{color,xspace}
\usepackage{subfigure}
\usepackage{balance}
\usepackage{cite}
\usepackage{color,soul}
\usepackage{mathtools}
\usepackage{makeidx}
\usepackage{robustindex}
\usepackage{float}
\usepackage{multirow}
\usepackage{textcomp}
\makeindex

\title{\LARGE \bf
Acting Is Seeing: Navigating Tight Space Using Flapping Wings
}

\author{Zhan Tu\textsuperscript{$1$}, Fan Fei\textsuperscript{$1$}, Jian Zhang, and Xinyan Deng
\thanks{\textsuperscript{$1$}These two authors contributed equally to this work.}
\thanks{The authors are with the School of Mechanical Engineering, Purdue University. (Email: xdeng@purdue.edu).}
}

\begin{document}

\maketitle
\thispagestyle{empty}
\pagestyle{empty}

\begin{abstract}
Wings of flying animals can not only generate lift and control torques but also can sense their surroundings. Such dual functions of sensing and actuation coupled in one element are particularly useful for small sized bio-inspired robotic flyers, whose weight, size, and power are under stringent constraint. In this work, we present the first flapping-wing robot using its flapping wings for environmental perception and navigation in tight space, without the need for any visual feedback. As the test platform, we introduce the Purdue Hummingbird, a flapping-wing robot with 17cm wingspan and 12 grams weight, with a pair of 30-40Hz flapping wings driven by only two actuators. By interpreting the wing loading feedback and its variations, the vehicle can detect the presence of environmental changes such as grounds, walls, stairs, obstacles and wind gust. The instantaneous wing loading can be obtained through the measurements and interpretation of the current feedback by the motors that actuate the wings. The effectiveness of the proposed approach is experimentally demonstrated on several challenging flight tasks without vision: terrain following, wall following and going through a narrow corridor. To ensure flight stability, a robust controller was designed for handling unforeseen disturbances during the flight. Sensing and navigating one's environment through actuator loading is a promising method for mobile robots, and it can serve as an alternative or complementary method to visual perception.
\end{abstract}
\section{Introduction} \label{sec:Intro} 
Bio-inspired Flapping Wing Micro Air Vehicles (FWMAVs) aim to mimic the stable and maneuverable flight of nature's flyers such as insects and hummingbirds. Autonomous navigation for such system remains challenging, especially in unstructured and tight spaces, due to the extremely stringent constraints on their weight, size, and power. These constraints, together with large vibrations induced from the high-frequency flapping wings, severely limit the available sensors that can be used for autonomous navigation.

In nature, the wings of flying animals are usually multi-functional: they not only can generate aerodynamic lift but also can provide somatosensory information to sense their surroundings. In this way, they play an important role in the neuromotor control to accommodate the uncertainties in their environments \cite{pringle1949excitation,dickinson1992directional,brown1993airflow,sterbing2011bat}. However, to this day, man-made flapping-wing vehicles generally use their wings only for actuation \cite{keennon2012development,ma2013controlled,phan2017design,zhang2017geometric,tijmons2018attitude,hines2015platform}. Employing dual functions of actuation and sensing in flapping wings like the ones in natural flyers has rarely been studied.

\begin{figure}[!t]
\begin{center}
\includegraphics[trim = 0mm 0mm 0mm 0mm, clip,width=0.9\columnwidth]{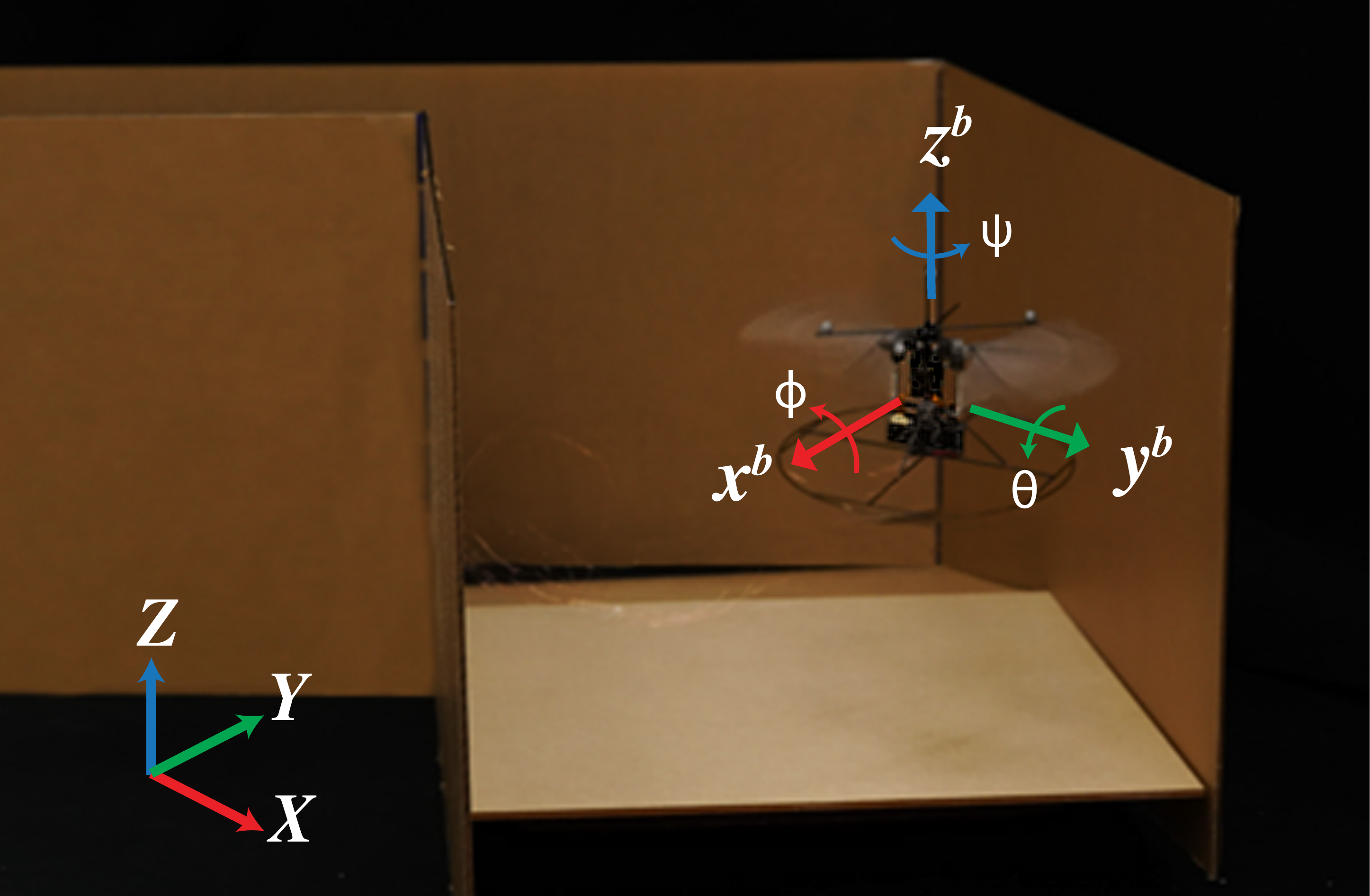}
\caption{
A motor-driven hummingbird robot was navigating a 1ft width corridor, guided by loading variations on the flapping wings. The platform has a wingspan of 17cm and weighs 12 grams. A 12cm diameter landing gear is attached for safe takeoff and landing.  By analyzing the current feedback from the motors, the terrain change and the wall barrier can be sensed by the reciprocating wings.
}
\vspace{-0.17in}
\label{fig:sysintro}
\end{center}
\end{figure}
In this paper, we present the first FWMAV-Purdue Hummingbird using flapping wing to sense the three-dimensional environment at run-time.
Our vehicle is designed to use the flapping wing actuation as a dual function element with sensing capabilities to enable autonomous navigation of the vehicle without visual feedback. Environmental changes, such as terrain, obstacles, and wind gust, induce instantaneous wings loading, which in turn manifest as changes in the feedback currents of the motors that actuate the wings. We can then filter and interpret this information to sense and map the environment, especially in compact spaces. To demonstrate the effectiveness, we tested our approach on a 12-gram hummingbird robot to work on several challenging tasks essential for navigation in tight spaces: following the varying terrain, following the wall, and passing through a narrow corridor with turns, without using any vision or standard proximity sensors. To ensure flight stability, we design a nonlinear robust controller to reject disturbances encountered during the tasks, since in tight spaces, aerodynamic forces on flapping wings are particularly noisy and complicated, and there will be some unforeseen contact dynamics as well. In general, discerning environmental variations from actuator loading is a promising sensing and navigation methodology for any type of mobile robots with stringent weight, size and power constraints, and can replace or complement other visual perception approaches.

\section{Related Work} \label{sec:Related work}

With the ever-increasing understanding of the flapping wing aerodynamics in recent years \cite{ellington1984aerodynamics, dickinson1999wing}, development of FWMAVs has made tremendous progress. Several flapping-wing platforms have successfully demonstrated stable hover flight \cite{keennon2012development,ma2013controlled,phan2017design,zhang2017geometric,tijmons2018attitude} and some interesting maneuvers \cite{chirarattananon2016perching,chen2017biologically,karasek2018tailless}. Yet, the flight performance is still unable to rival that of the natural flyers, especially when the environment is cluttered or unstructured. In fact, autonomous navigation in unexplored tight spaces remains an open question for mobile robots.

The sophisticated sensory system in nature's flyers is critical to their extraordinary environmental adaptability from reactive control to high-level navigation \cite{dickinson1992directional,sterbing2011bat}. Inspired by nature, in order to sense the surroundings and improve flight performance, numerous efforts have been devoted to onboard sensor design and implementation of FWMAVs. Among them, the visual sensors are the most widely used. At insect-scale, Harvard Robobee's ocelli-inspired optical sensor was able to stabilize the robot to its upright orientation \cite{fuller2014controlling}. At bird-scale, an onboard stereo vision system was successfully integrated on Delfly for obstacle avoidance \cite{de2014autonomous}. However, visual sensors typically have strict restrictions of light condition, and at such a small-scale, they show additional limitations due to weight, power and high computational load constraints. In this work, we present to using actuator loading for sensing, which can be an alternative or complementary method besides vision.

\begin{figure*}[htb]
\begin{center}
\includegraphics[trim = 0mm 0mm 0mm 0mm, clip,width=1.9\columnwidth]{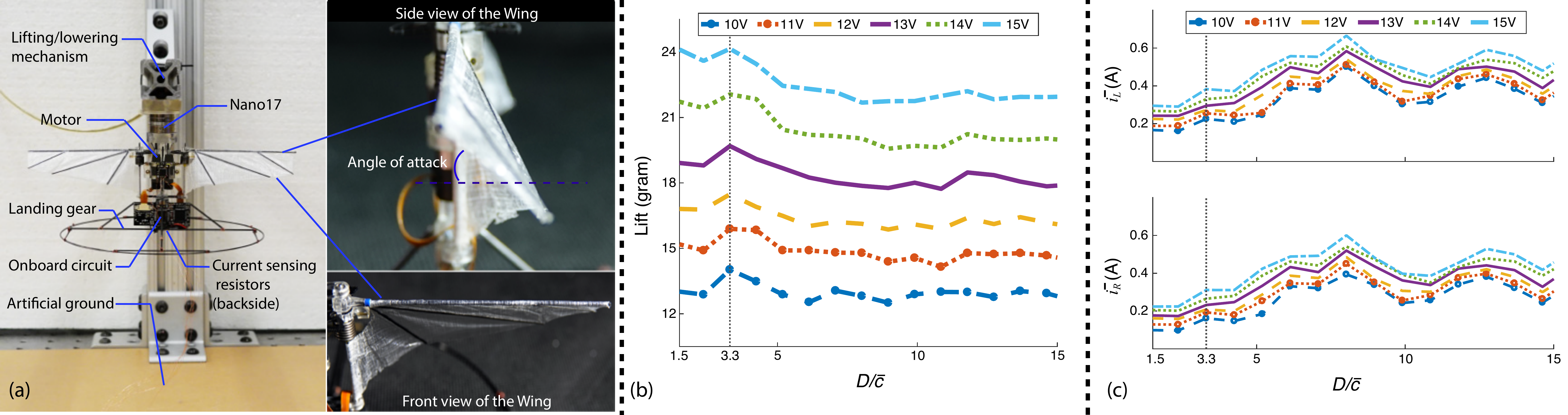}
\caption{
(a) Illustration of the setup for quantifying the ground effect. The details of the robot and the flexible wings are shown. The lifting/lowering mechanism is used for adjusting the clearance between the robot base and the ground. The robot is mounted on a 6-axis force/torque transducer and perpendicular to the artificial ground. The artificial ground is attached on a flat optic-table with horizontal calibration.
(b) The averaged lift measurement with ground effect is shown, where $D / \bar{c}$ represents the ratio of ground clearance and mean chord length of the wing.
(c) The dc bus current of the motor changes with the increased ground clearance, where $\bar{i_L}$ and $\bar{i_R}$ are cycle-averaged current feedback of the left and right wing respectively.
}
\label{fig:Fig2_ge}
\end{center}
\end{figure*}

Besides visual perception, haptic feedback provides another method for environment sensing. Inspired by the cockroaches antennas, researchers implemented artificial antennas on ground vehicles for tactile sensing \cite{lewinger2005insect, lee2008templates}. They demonstrated successful wall detection and following without visual cues. Similarly, as presented in \cite{giguere2006environment}, the legged robot can use contact responses of the leg to leverage a Bayesian classifier for terrain identification. The touch sensing strategy can be used on the FWMAVs as well. In this paper, we achieve the same function without using any specialized sensors, only measuring wing loading to infer the surroundings change. By taking advantages of the flexibility and reciprocating motion of the flapping wing, the safety of the FWMAV can be assured if the wing collided. In comparison, rigid-winged vehicles usually avoid hitting objects to prevent the wing wear and tear, e.g., drones need a cage-like shield to ensure passive safety when traveling through tight spaces with obstacles and turns \cite{thevoz2015flyability}.

\section{Environment Sensing Principle} \label{sec:Principle}
In this section, we first introduce our test platform: a motor-driven hummingbird robot. Then, we present the specific sensing principles of environmental conditions including ground perception, wall detection, and wind gust sensing.

\subsection{Platform Setup}\label{subsection:platform intro}
As shown in Fig.\ref{fig:sysintro} and Fig.\ref{fig:Fig2_ge}(a), our test platform is a bio-inspired hummingbird robot which has a wingspan of 17cm and weighs 12 grams. The wingbeat frequency is chosen at the resonant frequency of the system (34Hz) to improve power efficiency and maximize lift generation \cite{zhang2017design}. Two flexible wings are designed to achieve passive rotation with the optimal angle of attack (AoA) during flapping. Assuming a constant cycle-averaged fixed AoA, the aerodynamic force and drag can be modeled as a function of the angular speed of the wing motion ($\dot{\phi}_w$) \cite{ellington1984aerodynamics, dickinson1999wing}, which can be fully sensed by the motor current feedback as
\begin{equation} \label{eq:i_a calculation}
V_s - i_a R_a=K_a \dot{\phi}_w/N_g,
\end{equation}
where $V_s$ is the motor excitation voltage, $i_a$ is the armature current, $R_a$ is motor winding resistance, $K_a$ is motor torque constant, $\phi_w$ is the wing stroke angle, and $N_g$ is the gear ratio.

Two dc motors are used here to drive the wings independently with sinusoidal excitation. All essential thrust and control torques are generated by wing kinematics modulation \cite{doman2010wingbeat}. A customized onboard circuit is implemented for onboard actuation, sensing, and control. Two 0.4$\Omega$/0.5W sensing resistors (totally weights 0.03 grams) are used for motor current feedback.

\subsection{Terrain Sensing}\label{subsection:ground sense}
As the flapping wing flyer approaches the ground, the interaction between the wing-generated downwash airflow and the ground surface yields a strong variation of aerodynamic lift and drag forces \cite{kim2014hovering, lu2016ground}. This phenomenon is named ground effect, which potentially can be used for terrain change detection. 

For real hummingbirds, ground effect causes large lift enhancement. This phenomenon was presented by Kim et al via airflow visualization \cite{kim2014hovering}. Similar response was found on our bio-inspired hummingbird robot. In addition, we also observed a simultaneous reduction of the motor current on both wings while the vehicle is flying towards the ground, which indicates that the aerodynamic drags of the wings were affected by ground effect as well. Such correlation can be employed for terrain sensing and assist in altitude control.

The experiment setup is shown in Fig \ref{fig:Fig2_ge}(a). The robot is mounted on a 6-axis force/torque transducer (Nano17, ATI Ind. Automation) for lift and drag measurement. The onboard driver circuit provides the instantaneous current feedback of the motors. During the experiments, the vehicle is receding to the ground from 15$\bar{c}$ to 1.5$\bar{c}$ distance, where $\bar{c}=21.2mm$ is the mean chord length of the wing. On each test, the motor drive voltage changes from 10V to 15V with 1V stepsize. The result of a particular voltage is the average of total 20 datasets wherein one dataset logs 3 complete wingbeats data with the 2KHz sampling frequency. The results are presented in Fig.\ref{fig:Fig2_ge}(b) and Fig.\ref{fig:Fig2_ge}(c).

From Fig.\ref{fig:Fig2_ge}(b), the presence of ground effect is around 2-4$\bar{c}$. In this area, 2.3$\bar{c}$ and 3.3$\bar{c}$ are two unique points because they correspond the minimum drag and maximum lift, respectively. An interesting finding is that the most energy efficient flying altitude of our vehicle is about 3.3$\bar{c}$ since it maximizes the benefit from the ground effect with the optimal lift and near-optimal power efficiency. To quantify the results, the current feedback of 3.3$\bar{c}$ and the corresponding excitations is fitted as a linear function,
\begin{equation} \label{eq:fit_I-V}
\begin{aligned}
&i_{L_{3.3\bar{c}}} = 0.015 V_s + 0.062, &i_{R_{3.3\bar{c}}} = 0.014 V_s + 0.037.\\
\end{aligned}
\end{equation}
During the flight, we set 3.3$\bar{c}$ as a threshold of current feedback. Subsequently, a feedforward altitude controller can complement the main controller for terrain undulation adaptation and estimating the relative altitude.

\subsection{Wall Sensing}\label{subsection:wall sense}
Different from the ground effect study, the current feedback method cannot sense an obvious wall effect on our platform. It may be caused by the weak airflow along the spanwise direction of the wing. Similar results are found on both mechanical flapper and real hummingbird \cite{dickinson1999wing,kim2014hovering}. Therefore, to sense the wall, we allow the wings to perform collision detection. Unlike the conventional rigid-winged aircrafts, the flexibility and reciprocating motion of the flapping wings can naturally ensure the vehicle flight safety in collisions.

\begin{figure}[!t]
\begin{center}
\includegraphics[trim = 0mm 0mm 0mm 0mm, clip,width=0.85\columnwidth]{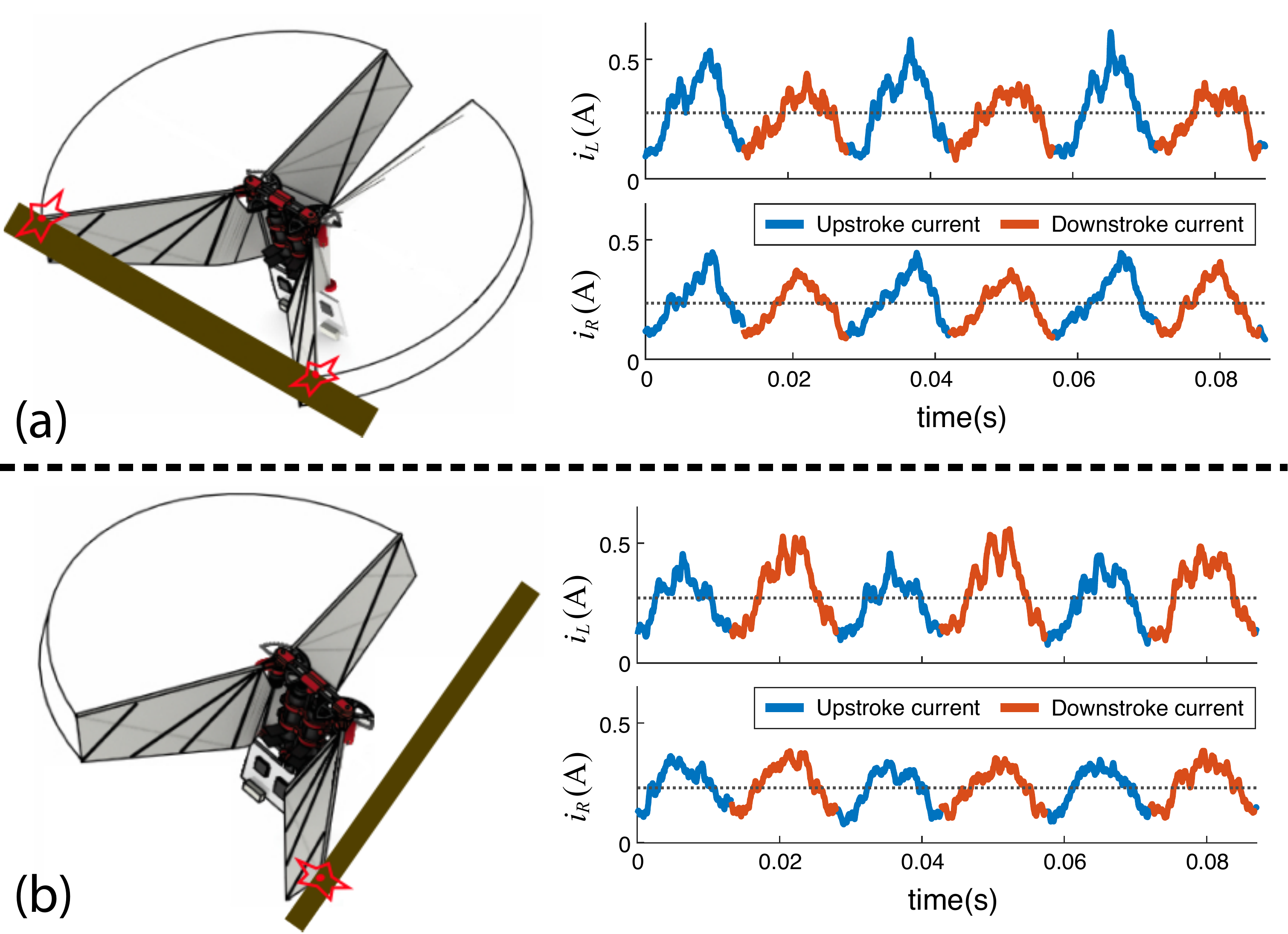}
\caption{
(a) When both wings collide the front-side obstacle in the forward flight (left figure), the instantaneous motor current $i_L$ and $i_R$ on both wings' upstroke movement are higher than normal (right figure).
(b) When the left wing's downstroke is bumping (left figure), only the $i_L$ downstroke current becomes greater (right figure).
}
\label{fig:wall_bump}
\vspace{-0.17in}
\end{center}
\end{figure}
From \cite{doman2010wingbeat}, for flapping flight, one wing beat consists of two phases, namely, upstroke and downstroke. Therefore, the half-stroke current difference between the two motors indicates the relative location of the obstacles. It is a unique advantage of flapping-wing vehicles that conventional aircraft don't have. In general, arbitrarily placed wall barriers can be categorized into two cases:
\begin{enumerate}

\item The wall is placed on the wingstroke direction to barricade the forward/backward flight path as shown in Fig.\ref{fig:wall_bump}(a). In a collision, both wings show increased current feedback on the upstroke whereas the other direction of the wing motion is unaffected. The characteristics of backward flight case can be derived in the same manner.

\item The wall is placed on the spanwise direction to barricade the left/right flight path as shown in Fig.\ref{fig:wall_bump}(b). In this case, only one wing will bump the obstacles, which can be used to distinguish case 1 and case 2.

\end{enumerate}
Therefore, combining case 1 and 2, there are a total of six directions can be sensed by two flapping wings.

For both wings, the current feedback on upstroke and downstroke direction need to be calibrated to deal with the fabrication imperfections. In our calibration, the ground clearance of the robot is set around $3.3\bar{c}$ for later flight test that involves the ground effect. The setup is same as that in the ground effect test besides an artificial wall for wing colliding. Throughout the test, the excitation still changes from 10-15V with 1V stepsize for each wing to quantify the current discrepancy between upstroke and downstroke.

\begin{figure}[!h]
\begin{center}
\includegraphics[trim = 0mm 0mm 0mm 0mm, clip,width=0.9\columnwidth]{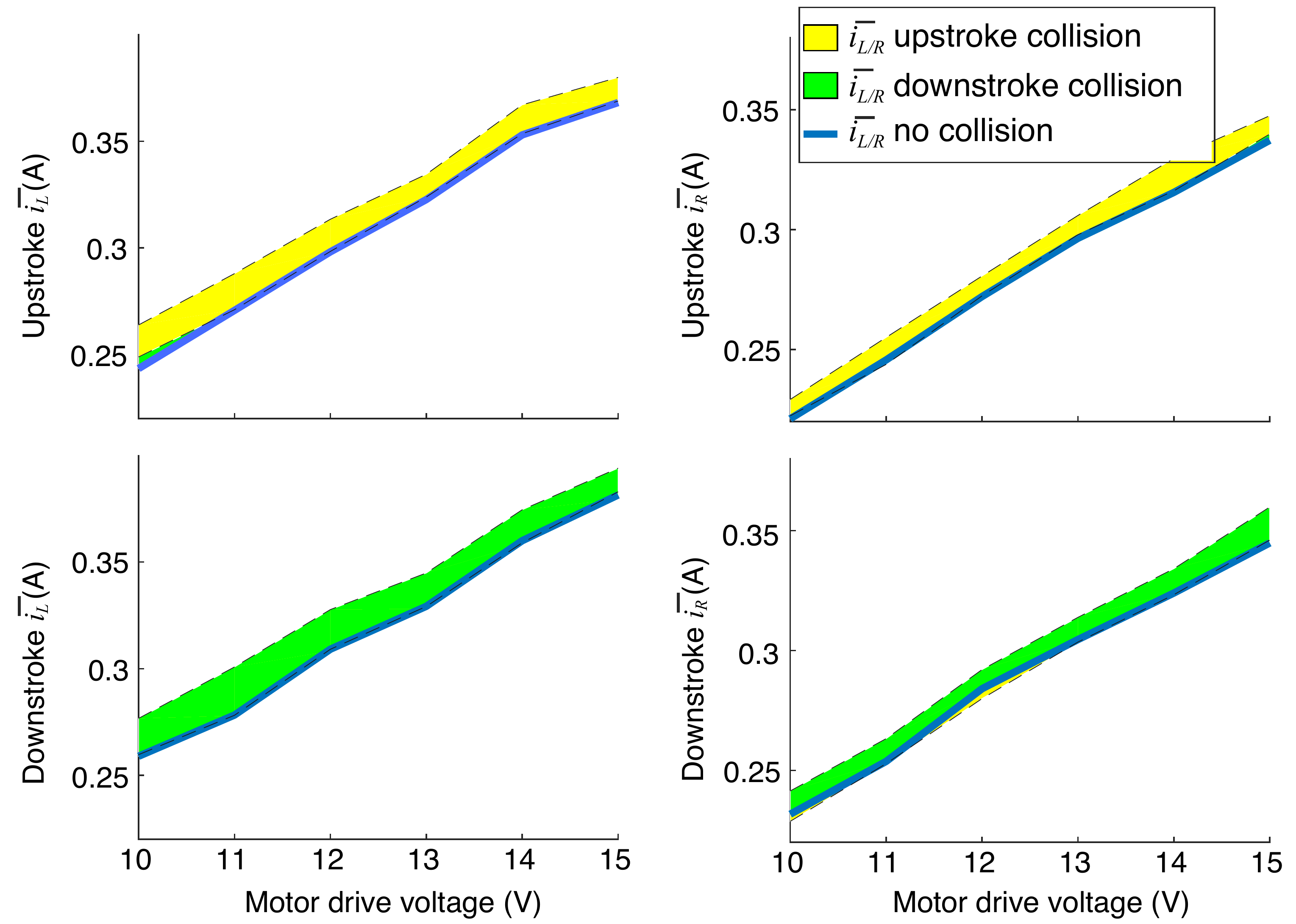}
\caption{
This calibration evaluates the averaged upper bound of the current feedback covering four different collision scenarios according to the instantaneous wing stoke direction.
}
\label{fig:wall_i_bound}
\vspace{-0.17in}
\end{center}
\end{figure}

The calibration result is shown in Fig.\ref{fig:wall_i_bound}. Both wings exhibit about 10\% increase of the averaged motor current when the wing bumps object. The collision-induced current variation rises linearly with the input voltage increases. We use this property to sense the wall. In flight test, if a collision was detected, the vehicle should perform evasive actions while recording the relative location of the obstacles.

\subsection{Wind Sensing}
Since the robot does not have any tactile receptors (birds' feather, bats' wing-hair or insects' wing pressure sensilla. etc.) over the wing surface, it cannot directly detect the airflow. Despite, when the robot is flying against the wind, wind gust can be sensed since the additional aerodynamic pressure is applied on the wings. In this situation, ideally, the response the motor current is the same as Fig.\ref{fig:wall_bump}(a), i.e., both wings will show increased upstroke current.

\section{Onboard Flight Control} \label{sec:Control}
\subsection{Robust Controller Design} \label{subsection:robust control}
In order to ensure stable flight in tight spaces, the noisy airflow of the ground effect and the instabilities caused by the collisions must be considered. To this end, we implement a robust controller to attenuate these disturbances. The controller is designed and fine-tuned in a simulation tool \cite{fei2019flappy}. The performance of the proposed controller is also validated and presented in a parallel study \cite{fei2019learning}.

With rigid body assumption, the vehicle's dynamics is described by
\begin{equation} \label{eq:body_dynamics1}
\begin{aligned}
&\dot{\bm{P}}=\bm{v},& &m\ddot{\bm{P}}=\bm{R}\bm{F}^b+m \bm{g},\\
&\dot{\bm{R}}=\bm{R} \hat{ \bm{\omega} },& &\bm{I} \dot{\bm{\omega}}^b = \bm{\tau}^b - \bm{\omega}^b \times \bm{I} \bm{\omega}^b,,
\end{aligned}
\end{equation}
where $\bm{P}$ and $\bm{v}$ are the position and velocity vectors of the vehicle in the inertial frame; $m$ is the total mass of the vehicle; $\bm{g}=[0,0,-9.8]^T$ is the gravity acceleration vector; $\bm{R}$ is the rotation matrix; $\bm{\hat{\cdot}}$ denotes the skew-symmetric matrix mapping from $\bm{\hat{a}b}$ to $\bm{a \times b}$; $\bm{\cdot}^b$ represents the body frame vector, including the 3-axis force $\bm{F}^b=[0,0,F_z]^T$, the vehicle angular velocity $\bm{\omega}^b$, and the 3-axis torque $\bm{\tau}^b=[\tau_x, \tau_y, \tau_z]^T$; $\bm{I}$ is the inertia matrix of the vehicle.

The thrust and body control torques are generated by the motor excitation modulations as
\begin{equation} \label{eq:force_mapping}
\begin{aligned}
&F_z = K_V V_s,& & \tau_x = K_\phi \delta V + \tau_{x_0}, \\
&\tau_y = K_\theta V_b + \tau_{y_0},& & \tau_z = K_\psi \sigma + \tau_{z_0}, \\
\end{aligned}
\end{equation}
where $F_z, \tau_x, \tau_y, \tau_z$ are thrust and 3-axis body torque; $K_V, K_\phi, K_\theta, K_\psi$ are lumped wing kinematics control gains; $V_s, \delta V, V_b, \sigma$ are motor drive voltage, differential voltage between two motors, voltage bias, and split-cycle parameter, respectively; $\tau_{x_0}, \tau_{y_0}, \tau_{z_0}$ are trim conditions of the vehicle which can be identified and calibrated by current feedback. 

A sliding surface $s_z$ for altitude control is defined as $s_z = \dot{e}_z + k_{s_z} e_z = \dot{z} - \dot{z_{eq}}$, where $e_z = z - z_r$ is the tracking error, $k_{s_z}$ is a positive gain and $z_{eq}$ is the equivalent tracking error. Similar to the altitude control, we define $\bm{s_\omega} = \bm{\omega} - \bm{\omega}_{eq}$ for lateral position and heading control, the detailed derivation of $\bm{\omega}_{eq}$ is presented and verified in \cite{chirarattananon2016perching}.

From above, system dynamics can be reconstructed by
\begin{equation} \label{eq:body_dynamics2}
\begin{aligned}
&m \dot{s}_z = K_z u_z + \bm{\Phi}_z^T \bm{\Theta}_z + d_z,\\
&\bm{I \dot{s}_\omega} = \bm{K_\omega} \bm{u_\omega} + \bm{\Phi_\omega^T \Theta_\omega} + \bm{d_\omega},\\
\end{aligned}
\end{equation}
where
\begin{equation} \label{eq:regressors}
\left\lbrace
\begin{aligned}
&K_z =  K_u cos\phi cos\theta,\\
&\bm{\Phi}_z^T=[-(g+\ddot{z}_{eq}), K_z u_0, 1],\\
&\bm{\Theta}_z = [m, 1, d_z]^T,  \\
&\bm{K_\omega} = diag([K_\phi, K_\theta, K_\psi]),\\
&\bm{\Phi}_\omega^T = [\bm{I_3}, -\bm{\text{diag}}([\bm{\omega} \times \bm{I \omega}+\bm{I \dot{\omega}_{eq}}]), \bm{I_3}],\\
&\bm{\Theta}_\omega = [\tau_{x_0}, \tau_{y_0}, \tau_{z_0}, 1, 1, 1, d_\phi, d_\theta, d_\psi]^T, \\
\end{aligned}
\right.
\end{equation}
the input $u_z=V_s$ and $\bm{u_w} = [\delta V, V_b, \sigma]$,
$\bm{\Phi}_z^T$ and $\bm{\Phi_\omega}^T$ are regressors,
$\bm{\Theta}_z$ and $\bm{\Theta_\omega}$ are model parameters,
$d_z$ and $\bm{d_\omega}$ denote lumped modeling errors and disturbances.

The control law is given by
\begin{equation} \label{eq:control_law}
\begin{aligned}
&K_z u_z = u_{z_m} + u_{z_l} +  u_{zr},\\
&\bm{K_\omega} \bm{u_\omega} = \bm{u_{\omega_m}} + \bm{u_{\omega_l}} +  \bm{u_{\omega_r}},
\end{aligned}
\end{equation}
where
\begin{equation} \label{eq:arc_terms}
\begin{aligned}
&u_{z_m} = -\bm{\Phi}_z^T \hat{\bm{\Theta}}_z,& &\bm{u_{\omega_m}} = -\bm{\Phi_\omega}^T \hat{\bm{\Theta}}_\omega,\\
&u_{z_l}  = -k_{z_l} s_z,& &\bm{u_{\omega_l}}  = -\bm{k}_{\omega_l} \bm{s}_\omega, \\
&u_{z_r}  = -\dfrac{h_z^2 s_z}{4 \epsilon_z}, & &\bm{u_{\omega_r}} = -\dfrac{[\bm{\text{diag}(h_\omega)}]^2 \bm{s_\omega}}{4 \bm{\epsilon_\omega}}, \\
\end{aligned}
\end{equation}
$u_{z_m}$ and $\bm{u_{\omega_m}}$ are model-based compensation terms, where $\bm{\hat{\Theta}_z}$ and $\hat{\bm{\Theta}}_\omega$ are from the reference model; $u_{z_l} $ and $\bm{u_{\omega_l}} $ are the linear stabilizing terms, where $k_{z_l}$ and $\bm{k}_{\omega_l}$ are positive gains; the formulation of the robust control terms $u_r$ and $\bm{u_{\omega_r}}$ are based on our previous study \cite{zhang2016instantaneous}, wherein $h_z$ and $\bm{h_\omega}$ are the upper bound of the modeling uncertainties, $\epsilon_z$ and $\bm{\epsilon_\omega}$ represent the approximation accuracy between the ideal sliding mode control and the proposed method.

The stability proof is similar to \cite{zhang2016instantaneous} with different input. 

\subsection{Navigation with Current Feedback} \label{subsection:navigation}
As discussed in section \ref{sec:Principle}, the motor current feedback of our FWMAV can be used to identify different environmental factors and further assist in navigation, especially when the robot is flying in compact spaces without vision or proximity sensors.

\begin{figure*}
\begin{center}
\includegraphics[trim = 0mm 0mm 0mm 0mm, clip,width=1.93\columnwidth]{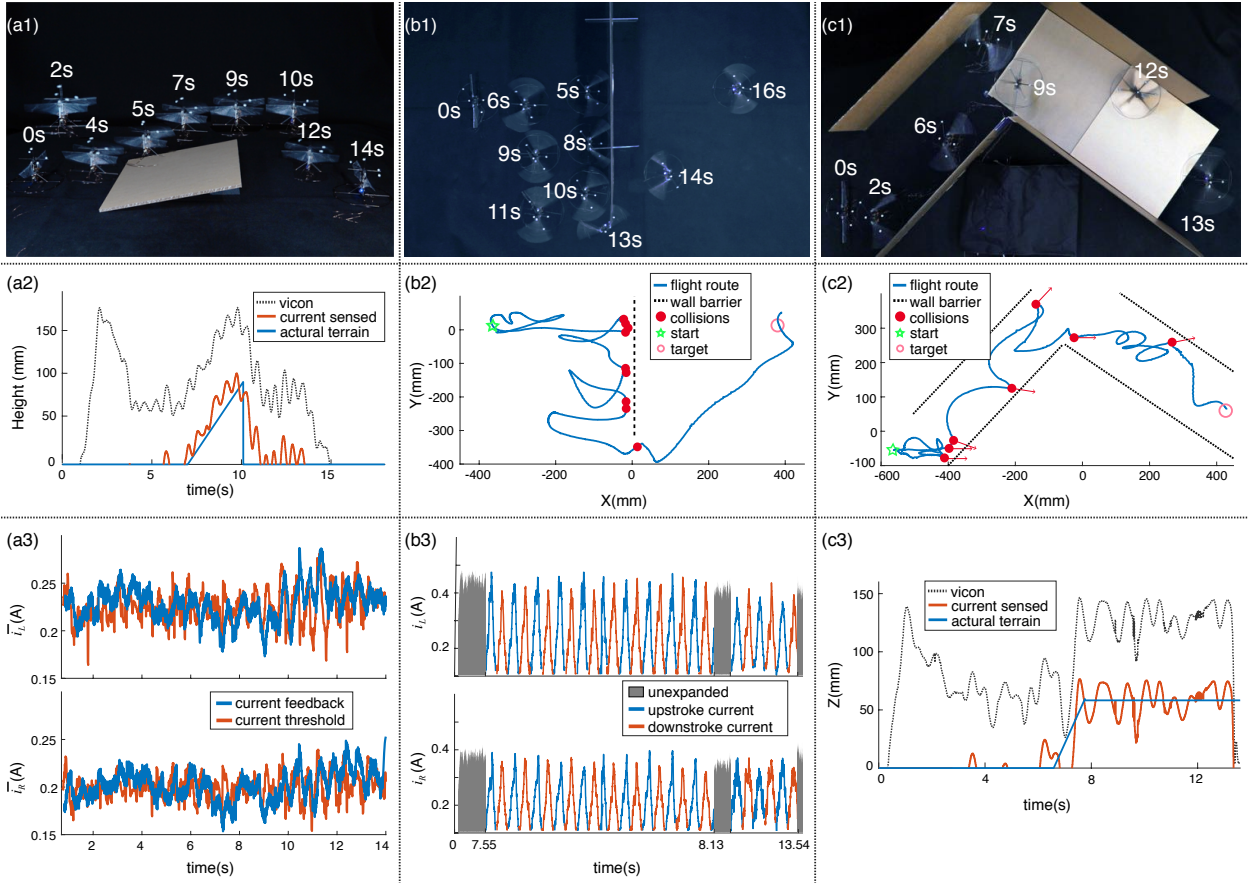}
\caption{
(a1) Compound time sequences of the terrain following and mapping test. The vehicle was planned to move point-to-point. A ramp was placed to change the terrain. The robot successfully cross over the ramp with ground effect sensing. The ground effect had been detected at around 2-4s, 7-8s, 9-10s, and 11-13s.
(a2) Altitude sensing result through ground effect. The terrain mapping result is depicted accordingly.
(a3) The current thresholds ($i_{L_{3.3\bar{c}}}$ and $i_{R_{3.3\bar{c}}}$) are calculated by the input voltage. The cycle-averaged current feedback is captured from the motors at run-time.
(b1) Compound time sequences of the wall barrier sensing and following test. The vehicle was planned to move point-to-point. A wall was placed to barricade the shortcut path. The robot detect the wall by wing collisions. As the wing bumps, the flight route is altered for collision avoidance. After a few route alterations (at around 5s, 8s, 10s, 13s in this flight) the vehicle will pass the wall. The outline of the obstacles can be mapped simultaneously.
(b2) The default flight mission is to move from the starting point (-350,0) to the endpoint (350,0). Since a wall barricades the shortcut path (go straight) the robot needs to pass the wall by using the strategy presented in section \ref{subsection:navigation}. With a couple of collision detections and path alternations, the final actual flight path is shown.
(b3) The instantaneous current measurements corresponding to this flight. The two enlarged areas with detailed upstroke/downstroke current feedback are corresponding to the 8s and 13s collisions. They demonstrate case 1 and 2 respectively as mentioned in section \ref{subsection:wall sense}. 
(c1) Compound time sequences of the hummingbird robot passing through a corridor. The vehicle aims to arrive its destination by combining both ground sensing and wall sensing since the corridor is designed with a turn and changed terrain. Similar to the above two test, with the recorded ground surface information and collision coordinates, the contours of the obstacles can be extracted.
(c2) The actual route is determined by the somatosensory-like feedback of the reciprocating wings. The red dot and arrow indicate the perceived obstacle point and the corresponding heading direction at the moment of bumping.
(c3) Same as (a2).
}
\label{fig:result_all_in_one}
\vspace{-0.17in}
\end{center}
\end{figure*}

Based on section \ref{subsection:ground sense}, a feedforward command $\delta z_r = K_{\hat{z}} (i_{L/R} - i_{{3.3\bar{c}}_{L/R}})$ representing the relative altitude variation was sent to the altitude controller $z_r = z_r + \delta z_r$ to follow the undulated terrain, where $K_{\hat{z}}$ is a positive proportion gain. Meanwhile, the terrain change is recorded for mapping purpose. Note, this sensing strategy requires roll, pitch control error to be sufficiently small. 

To suppress the undesired z-axis oscillation caused by the noise of the current measurement, a low-pass filter with cut-off frequency on 200Hz is adopted for signal conditioning. Moreover, a dead-band of the current feedback is set to smooth the altitude movement, which is tuned to be 75\%-100\% of $i_{L/R_{3.3\bar{c}}}$ on our test platform. In the dead-band, the vehicle does not respond to the feedforward control. Although the dead-band setting sacrifices measurement accuracy, it damps z-axis oscillation effectively and improves flight stability.

On the X-Y directions, ideally, the robot is located by the position sensor. With the proposed obstacle detection method, the vehicle is ready to alter the route for obstacle avoidance. In actual flight, the position sensor is not always working properly, which causes that the vehicle cannot locate itself and plan the route. To address this issue, we temporarily use the onboard inertia sensor (IMU) to approximate the position information. Since the altitude is still controlled by current feedback, the transformation matrix is defined as
\begin{equation} \label{eq:position_trans}
\begin{bmatrix}
x_k \\
y_k \\
1 \\
\end{bmatrix} =
\begin{bmatrix}
\begin{bmatrix} cos\psi & -sin\psi \\ sin\psi & cos\psi \end{bmatrix} & [x_{k-1}, y_{k-1}]^T \\
 0 & 1 \\
\end{bmatrix} \bm{p^b}
\end{equation}
where $\bm{p^b} = (x^b, y^b,1)$ is the augmented lateral position vector in the body frame.

From section \ref{subsection:wall sense}, the relative location about vehicle and obstacles can be diagnosed. While the obstacles were detected, our robot was programmed to avoid it with a simple but robust method: retreating along the opposite direction of the detected obstacle, and then altering the original flight trajectory for avoidance. We specify that the route shift is always perpendicular to the current moving direction. If the robot bypassed the bumping location, it will be back to the default route. 

\section{Experimental Results and Discussion} \label{sec:Experiments}
In order to validate the proposed method, firstly, the hummingbird robot was programmed to follow the terrain and the wall barrier, respectively. Afterward, the robot performed to blindly go through a corridor with barrier blocking and terrain change together. These experiments successfully demonstrated that the flapping-wing vehicles are able to sense and map the surroundings by using their wings as the primary sensor. In the experiments, the ground truth was provided by a VICON motion tracking system (http://www.vicon.com). An onboard real-time sensor fusion algorithm \cite{tu2018realtime} was implemented for attitude stabilization. Position feedback was a fused signal which combines VICON and onboard accelerometer measurements.

\subsection{Terrain Following}
In this test, a ramp with 1ft length and $20^\circ$ slope is placed on the flying path. The robot aims to cross over it blindly. At the beginning of the experiment, the robot ascends to a relatively high altitude, and then descends while the robot is searching the ground surface. If the robot ensured that the ground has been detected, it starts to move forward at a constant speed. When the ground clearance deviates out of the dead-band zone of the predefined threshold, the feedforward control command will be involved for accommodating the terrain fluctuations.

Fig.\ref{fig:result_all_in_one}(a1) is a time sequential result. The corresponding current feedback data is shown in Fig.\ref{fig:result_all_in_one}(a3). With the calculated thresholds and the real-time current feedback of the two motors, the actual ground clearance of the vehicle can be approximately sensed, e.g., when the vehicle is searching the ground at the very beginning (around 2-4s), the current feedback $\bar{i_{L}}$ and $\bar{i_{R}}$ are both beyond the threshold which indicates the ground clearance is too large and needs the controller to rectify. Accordingly, the following terrain changes can be deduced as well. As the robot navigating the ramp, it continuously estimates the terrain changes. The mapping result is shown in Fig.\ref{fig:result_all_in_one}(a2), the error interval of the  estimation result can be maintained at two centimeters around. As seen from this result, the hummingbird robot is capable of sensing altitude using its wings.

\subsection{Wall Following}
In this test, the robot is performed to point-to-point flight. A 1$\times2$ft$^2$ wall barrier is placed to block the desired flying path between the start and end points. The robot relies on the wing collisions to detect the presence of the obstacles, and then alters the flight trajectory until it passed the obstacles.

Fig.\ref{fig:result_all_in_one}(b1) is a time sequential result. The actual flight route and sensed collisions are shown in Fig.\ref{fig:result_all_in_one}(b2). Through the proposed methodology in section \ref{subsection:wall sense}, the obstacles can be detected by the wing bumping. In order to eliminate false positives, multiple collisions at a certain point are allowed. The overlayed collision points in Fig.\ref{fig:result_all_in_one}(b2) is the sensing result, which guides the vehicle to bypass the wall. With a couple of violent bumps, the robots can still maintain stable flight, which demonstrates the robustness of the flapping-wing vehicles.

\subsection{Passing A Narrow Corridor} 
The challenge of this experiment is not a simple merge of the different environmental factors. In this test, the corridor width is only 1ft, which is less than two wingspans, and in some spots, it loses position feedback. Without any additional sensors and aiding information, flying and exploring in a tight, unexplored, obstacle-filled space can be a big challenge for most MAVs. The success of this experiment demonstrates that flapping-wing vehicles have a natural advantage in the perception and adaptation of complex environments.

\begin{figure}[!h]
\begin{center}
\includegraphics[trim = 0mm 0mm 0mm 0mm, clip,width=0.9\columnwidth]{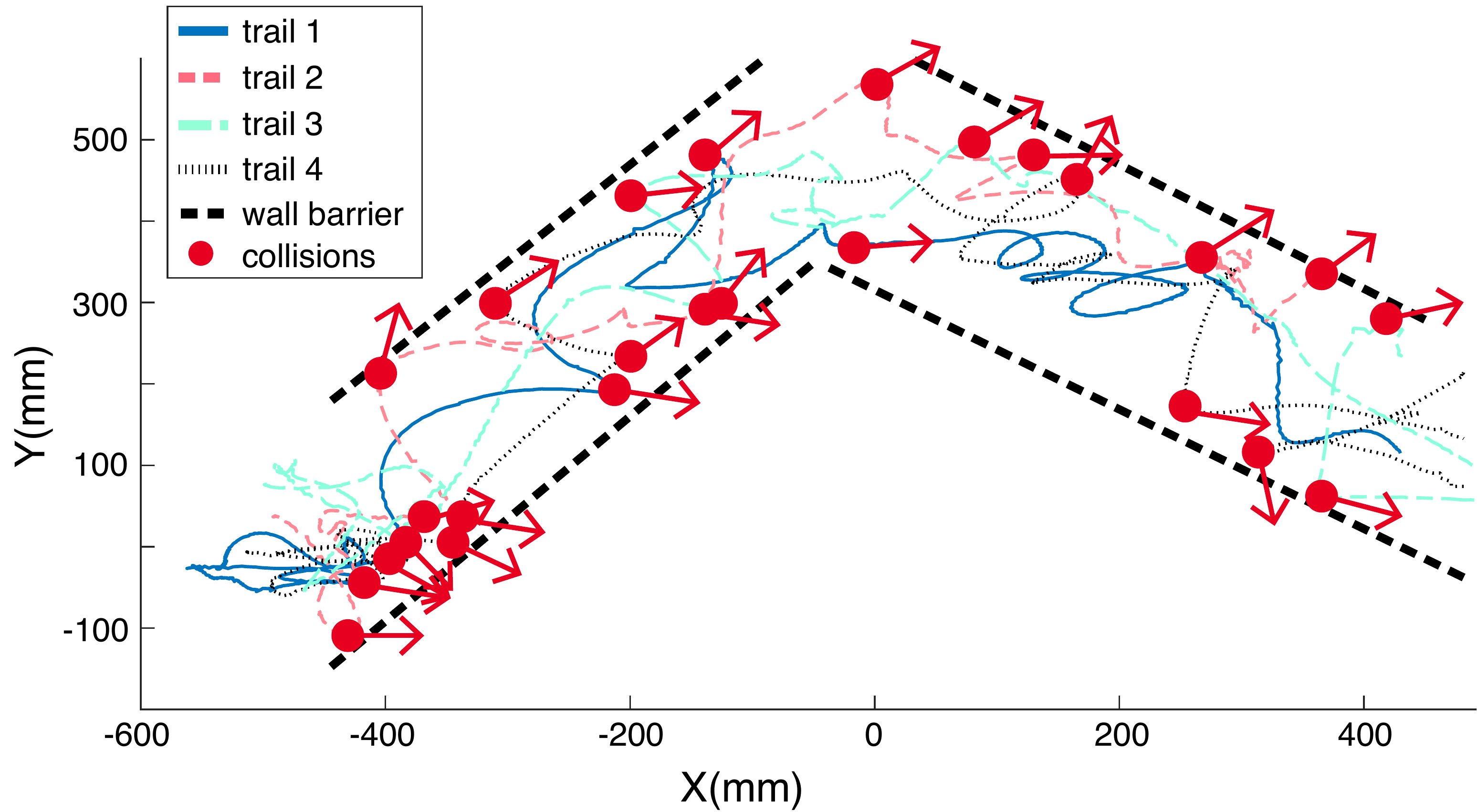}
\caption{After four flights, the contours of the corridor is clearly shown.}
\label{fig:4trails}
\vspace{-0.17in}
\end{center}
\end{figure}

Fig.\ref{fig:result_all_in_one}(c1) is a time sequential result. The actual flight path and the detected obstacles are shown in Fig.\ref{fig:result_all_in_one}(c2) and Fig.\ref{fig:result_all_in_one}(c3). The result shows that without visual sensor support, the robot successfully finished this navigation task with solely the interpreted environmental information from reciprocating wings. Furthermore, after multiple flight trails, the distribution of collision points can be more uniform, which results in a more complete and accurate outline mapping of the corridor interior as shown in Fig.\ref{fig:4trails}.

\section{Conclusions}
Inspired by flying animals, flapping-wing vehicles have their unique advantage of using the reciprocating wings not only as actuators but also as a somatosensory sensor to provide the information of their surroundings. Combining sensing and actuation within a single apparatus can gain many benefits, such as no additional payload, relatively high bandwidth and sensitivity, and low computation load, which hold a great promise on a variety of applications for the weight and size constrained robots. In this work, we presented the first FWMAV-Purdue Hummingbird with the employment of actuator loading to sense and aid autonomous navigation in tight spaces. We have shown that this hummingbird robot can obtain the somatosensory-like feedback by employing the instantaneous motor current feedback of the wing system. The feasibility of utilizing it as the primary sensor for traveling through a tight, unexplored, obstacle-filled space has been demonstrated experimentally. The fundamental sensing principle can be generalized to any mobile robots with stringent weight, size, and power constraints, serving as an alternative or complementary method to other sensing approaches. 
In future work, besides the wing loading information, we will incorporate more available onboard sensors, such as IMUs and  cameras, to facilitate robust localization and navigation in more complex environments.

\balance

\bibliography{cite_nav}
\bibliographystyle{IEEEtran}

\end{document}